\documentclass{article} 
\usepackage{tencent_ailab_tech_report}
\usepackage[colorlinks = true,
            linkcolor = blue,
            urlcolor  = blue,
            citecolor = blue,
            anchorcolor = blue]{hyperref}

\usepackage{tcolorbox} 
\newtcolorbox{alprompt}[1]{
        boxrule = 1pt,
        fontupper = \small\tt,
        fonttitle = \bf\color{black},
        arc = 2pt,
        rounded corners,
        colframe = black,
        colbacktitle = white!97!yellow,
        colback = white!97!yellow,
        title = #1,
}

\usepackage{microtype}
\usepackage{hyperref}
\usepackage{url}
\usepackage{booktabs}
\usepackage{enumitem}
\usepackage{multicol}
\usepackage{multirow}
\usepackage{CJKutf8}
\usepackage{amsmath}
\usepackage{siunitx}
\usepackage{floatflt}
\usepackage{wrapfig}
\usepackage{graphicx}
\usepackage{booktabs}
\usepackage{wrapfig}
\usepackage{authblk}
\usepackage{lipsum}

\usepackage{algorithm}
\usepackage{algorithmicx}
\usepackage{algpseudocode}
\usepackage{microtype}
\usepackage{graphicx}
\usepackage{multirow}
\usepackage{booktabs} 
\usepackage{pifont}  
\usepackage{graphicx}  
\usepackage{subcaption} 
\usepackage{hyperref}

\usepackage{amssymb} 

\algnewcommand{\LeftComment}[1]{\Statex \(\triangleright\) #1}

\usepackage{array}
\usepackage{amsmath}
\usepackage{amssymb}
\usepackage{mathtools}
\usepackage{amsthm}
\usepackage{arydshln}
\usepackage[capitalize,noabbrev]{cleveref}
\usepackage{adjustbox} 
\usepackage{enumitem}

\theoremstyle{plain}

\theoremstyle{definition}

\theoremstyle{remark}

\usepackage[textsize=tiny]{todonotes}

\definecolor{nred}{RGB}{196, 38, 11}
\definecolor{ngreen}{RGB}{18, 141, 21}
\definecolor{nblue}{RGB}{41, 52, 190}
\definecolor{hzw}{RGB}{223, 97, 76}
\definecolor{lt}{RGB}{54, 89, 170}

\newcommand{\ignore}[1]{}

\colmfinalcopy

\newcommand{\method}[0]{\textsc{Panel}}

\title{Dancing with Critiques: Enhancing LLM Reasoning with \\Stepwise Natural Language Self-Critique}

\author[ ]{Yansi Li\thanks{Equal Contribution. Work was done when Yansi Li, Xingyu Chen and Zhiwei He were interning at Tencent.}~~$^{,1,2}$}
\author[ ]{Jiahao Xu$^{*,1}$}
\author[ ]{Tian Liang$^{*,1}$}
\author[ ]{Xingyu Chen$^{1,2}$}
\author[ ]{Zhiwei He$^{1,2}$}
\author[ ]{\mbox{Qiuzhi Liu}$^{1}$}
\author[ ]{\mbox{Rui Wang}$^{2}$}
\author[ ]{Zhuosheng Zhang$^{\dag, 2}$}
\author[ ]{\mbox{Zhaopeng Tu}\thanks{Correspondence to: Zhaopeng Tu \textless zptu@tencent.com\textgreater~and Zhuosheng Zhang \textless zhangzs@sjtu.edu.cn\textgreater.}~~$^{1}$}
\author[ ]{\mbox{Haitao Mi}$^{1}$}
\author[ ]{Dong Yu$^{1}$}

\affil[1]{Tencent}
\affil[2]{Shanghai Jiao Tong University}

\begin{document}

\maketitle

\begin{figure}[h]
    \centering
    \vspace{-20pt}
    \includegraphics[width=\linewidth]{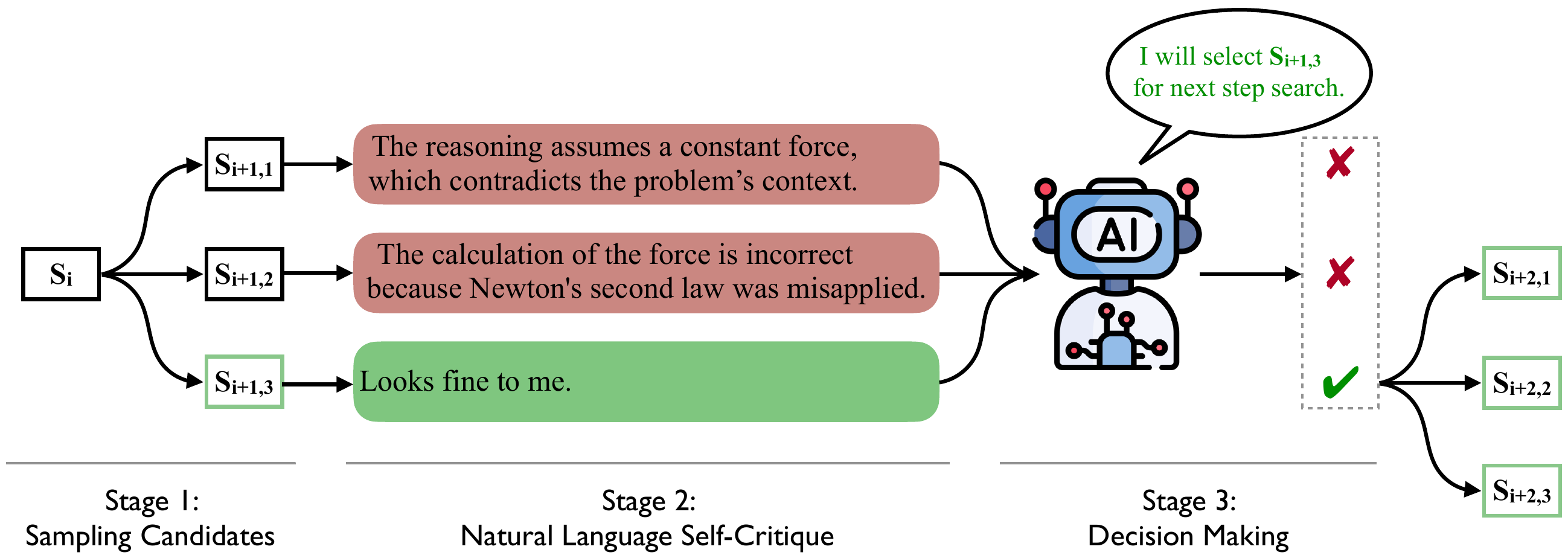}
    \caption{An illustration of our proposed \method, a novel inference time scaling framework that incorporates rich natural language self-critique to guide step-level search in reasoning tasks, moving beyond traditional scalar correctness scores. 
    Rather than relying solely on scalar outputs from task-specific verifiers, \method employs natural language feedback to offer nuanced insights into each reasoning step's strengths and weaknesses. Furthermore, \method~dynamically selects the best candidates, representing a fundamental departure from conventional verifier-based strategies that always choose the candidates with the highest PRM score.}
    \label{fig:framework}
    \vspace{20pt}
\end{figure}

\begin{abstract}
Enhancing the reasoning capabilities of large language models (LLMs), particularly for complex tasks requiring multi-step logical deductions, remains a significant challenge. Traditional inference time scaling methods utilize scalar reward signals from process reward models to evaluate candidate reasoning steps, but these scalar rewards lack the nuanced qualitative information essential for understanding and justifying each step. In this paper, we propose a novel inference-time scaling approach -- stepwise natural language self-critique (\method), which employs self-generated natural language critiques as feedback to guide the step-level search process. By generating rich, human-readable critiques for each candidate reasoning step, \method~retains essential qualitative information, facilitating better-informed decision-making during inference. This approach bypasses the need for task-specific verifiers and the associated training overhead, making it broadly applicable across diverse tasks. Experimental results on challenging reasoning benchmarks, including AIME and GPQA, demonstrate that \method~significantly enhances reasoning performance, outperforming traditional scalar reward-based methods. Our code is available at \url{https://github.com/puddingyeah/PANEL} to support and encourage future research in this promising field.
\end{abstract}

\section{Introduction}

Large language models (LLMs) have significantly transformed natural language processing by enabling sophisticated reasoning and problem-solving abilities. However, enhancing the reasoning capabilities of LLMs, especially in complex tasks that require multi-step logical deductions, remains a significant challenge. One critical technique for addressing this challenge is {\bf inference time scaling}, which strategically allocates computational resources during inference to explore a broader space of potential reasoning paths beyond single, deterministic trajectories. Recent methods employing inference time scaling have demonstrated the effectiveness of this strategy in improving the robustness and accuracy of LLMs' reasoning processes \citep{yao2024tree, OpenAI_2024, snell2024scaling, guo2025deepseek, Team_2024a}.

A prominent framework for implementing inference time scaling is \textbf{step-level tree search}, which iteratively explores possible reasoning steps to construct a solution path \citep{Villalobos_Atkinson_2023, luo2024improve, feng2023alphazero}. Central to this framework is the mechanism for evaluating and selecting the most promising reasoning paths. Traditional approaches assess the quality of each candidate step using step-level verifiers, which often utilize scalar reward signals derived from process reward models (PRMs) \citep{lightman2024lets, wang2024math}. These verifiers output numerical scores representing the correctness or desirability of steps, guiding the search algorithm towards paths with higher scores.

However, relying on scalar rewards introduces significant limitations. \textbf{First}, reducing complex reasoning steps to single numerical values inevitably \emph{sacrifices nuanced qualitative information essential for understanding and justifying each step}. Important insights, justifications, and potential errors may be overlooked, hindering the model's ability to perform complex reasoning. \textbf{Second}, effective verifiers are often \emph{task-specific and require substantial training on annotated datasets} that may not be available for many advanced reasoning tasks, particularly in STEM domains. \textbf{Finally}, the development and integration of these verifiers \emph{impose considerable computational overhead and complexity} \citep{guo2025deepseek}.

In this paper, we present a novel inference-time scaling approach called \textbf{stepwise natural language self-critique} (\method). Instead of relying on scalar reward signals from an external verifier, \method~employs self-generated natural language (NL) critiques as a feedback mechanism to guide the step-level tree search process. By generating rich, human-readable critiques for each candidate reasoning step, the model retains the qualitative information necessary for comprehensive understanding and justification. This approach offers several key advantages:
\begin{enumerate}[leftmargin=10pt]
    \item NL critiques provide detailed explanations of the strengths and weaknesses of each reasoning step, facilitating better-informed decision-making during the search process.
    \item Unlike task-specific verifiers, NL critiques can be generated by the policy model itself across diverse tasks without requiring specialized training data. This makes \method~suitable for a wide range of complex reasoning problems.
    \item By reusing the policy model to generate critiques, \method~circumvents the considerable overhead associated with training dedicated verifiers, streamlining the inference process.
\end{enumerate}

The remainder of this paper is organized as follows: Section~\ref{sec:method} details the proposed \method~framework, elaborating on the integration of NL critique within the step-level search algorithm and the mechanisms for leveraging critique feedback to guide the search process. Section~\ref{sec:experiment} presents a comprehensive empirical evaluation of \method~across a range of challenging reasoning tasks, demonstrating its effectiveness and advantages over existing approaches. We discuss related work in Section~\ref{sec:related_work} and conclude with our findings in Section~\ref{sec:conclusion}.

Our main contributions are as follows:
\begin{enumerate}[leftmargin=10pt]
    \item We propose \method, a novel inference time scaling framework that incorporates rich natural language self-critique to guide step-level search in reasoning tasks, moving beyond traditional scalar correctness scores.
    \item We provide a comprehensive analysis demonstrating how \method~addresses the limitations of existing scalar verifiers, offering a more informative, versatile, and efficient approach applicable to diverse reasoning tasks.
    \item We conduct extensive experiments on challenging reasoning benchmarks to validate the effectiveness of \method, showcasing significant improvements in reasoning performance by leveraging nuanced NL feedback.
\end{enumerate}

\section{\method}
\label{sec:method}

This section provides an overview of \method, our novel framework designed to enhance LLMs reasoning capabilities. \method~innovatively integrates natural language (NL) critique as a feedback mechanism directly into a step-level search process. Our core motivation stems from the recognition of natural language as a universal and robust feedback signal~\cite{ke2023critiquellm}, uniquely suited to address the diverse challenges of complex reasoning tasks across various domains.  

\subsection{\method~Framework}
We introduce \method, the first strategy to introduce natural language critique into the step-wise search algorithm, and validate its effectiveness in LLM reasoning across not only mathematical reasoning tasks but also various STEM tasks.

\paragraph{Stage1: Sampling Candidates}
The initial phase of \method~mirrors the candidate expansion phase in conventional step-level search algorithms. To effectively balance both certainty and diversity in our candidate pool, we employ a dual sampling strategy. Firstly, to capture more certain and likely next steps, we utilize greedy decoding. This approach selects the highest probability token at each decoding step, aiming to generate reasoning steps that the LLM deems most probable. Secondly, to introduce diversity and explore a broader range of potential reasoning pathways, we complement greedy decoding with random sampling with temperature\footnote{Temperate is 0.6 across all the experiments.}. 
This combination ensures that our candidate set encompasses both highly probable and more exploratory directions for the subsequent stages of the \method~framework to evaluate.

\paragraph{Stage2: Natural Language Self-Critique}
In the second stage, the \method~framework harnesses the expressive power of NL self-critique to evaluate the quality of each candidate's reasoning step generated in Stage 1.  Critically, unlike conventional approaches relying on scalar metrics, this stage leverages natural language critique to provide nuanced and human-interpretable justifications for the strengths and weaknesses of each candidate.  

This is particularly significant because natural language critique can be inherently \textbf{task-specific}, drawing upon relevant domain knowledge and contextual understanding that is often inaccessible to fixed scalar evaluation strategies. For instance, as demonstrated in Figure~\ref{fig:framework}, NL self-critique effectively pinpoints reasoning errors in candidate $\textbf{S}_{i+1,1}$ where errors stemming from \textit{incorrect assumptions} and candidate $\textbf{S}_{i+1,2}$ where the \textit{Newton's second law is misapplied}.  Both of these instances of incorrect reasoning are explicitly specific to the \textbf{Physics} domain, showcasing the capacity of NL critique to provide feedback grounded in the nuances of task-relevant knowledge, a capability absent in conventional scalar-based evaluation.

Critically, NL critique offers a robust signal that is broadly applicable, moving beyond the limitations of task-specific verifiers. To ensure this robustness, our NL critique is carefully designed to be effective across a wide range of complex reasoning tasks within the STEM domain, from mathematical problem-solving to physics and beyond (see Appendix~\ref{sec:nl_critique_template}). This stage facilitates a more comprehensive and adaptable assessment of reasoning step quality with natural language critiques.

\paragraph{Stage3: Decision Making}
The final stage of the \method~is decision Making, which deviates significantly from conventional step-level search algorithms. Instead of relying on a pre-defined scalar metric or external verifier to directly select the candidate with the maximal score, \method~dynamically leverages the LLM itself to make the selection.  As demonstrated in Figure~\ref{fig:framework}, the policy model analyzes the nuanced feedback associated with each candidate's reasoning step. Then the policy model assesses the overall quality and potential of each candidate and selects the candidate $\textbf{S}_{i+1,3}$ based on its justification with the NL critiques. Consequently, \method~transcends simple scalar-based selection by enabling a more sophisticated and context-aware decision process, allowing the framework to dynamically choose the most promising candidate to extend the reasoning trace.

Finally, once the candidate is selected, a new iteration of searching steps would start until reaching the final answer. Through this process, \method~introduces the NL critique as the feedback signal into step-level searching algorithms.

\subsection{Formulation}
This subsection details how \method~ leverages Natural Language (NL) self-critiques to refine step-level tree search. Unlike conventional methods that maximize scalar rewards, \method~ aims to identify step sequences justified by favorable NL critiques.

We begin by contrasting with the standard objective of value-based search algorithms. Given a question prompt $x$, a policy model $\theta$, and a PRM verifier $V_\phi$, the traditional goal is to maximize the expected cumulative reward:
\begin{equation}
\label{eq:objective_scalar_reward}
\underset{s_1, ..., s_N}{\operatorname{argmax}} \sum_{t=1}^{N} \mathbb{E}_{s_t\sim p(\cdot | s_{<t}, x; \theta)}[V_\phi(s_t| s_{<t})],
\end{equation}
where $s_t$ represents the state at step $t$, and $V_\phi(s_t|s_{<t})$ is the scalar reward from the PRM verifier for transitioning to $s_t$.

\paragraph{NL Critique as Step-Level Justification}
In \method, instead of scalar rewards, we use NL critiques to provide richer step-level justifications. For each step $t$, the policy model $\theta$ generates a set of candidate steps $\mathcal{C}_t = \{c_{t,1}, ..., c_{t,K}\}$.  A critique model $Q_\psi$ then produces corresponding NL critiques $\mathcal{Q}_t = \{q_{t,1}, ..., q_{t,K}\}$ for each candidate:
\begin{equation}
\mathcal{Q}_t = \{Q_\psi(c_{t,k}|s_{<t}) | c_{t,k} \in \mathcal{C}_t \}.
\end{equation}

\paragraph{Critique-Driven Step Selection}
The core of \method~ lies in using these critiques to guide step selection.  The policy model $\theta$ is designed to process both candidate steps $\mathcal{C}_t$ and their critiques $\mathcal{Q}_t$.  The probability of selecting a candidate step $c_{t,k}$ at step $t$ is conditioned on both sets, alongside historical context $s_{<t}$:
\begin{equation}
p(A_k | s_{<t}, \mathcal{C}_t, \mathcal{Q}_t; \theta).
\end{equation}

Effectively, instead of directly maximizing a scalar reward, \method~ aims to find a step sequence associated with a globally "favorable" set of NL critiques $\mathcal{Q} = \{\mathcal{Q}_1, ..., \mathcal{Q}_N\}$.  This can be conceptually represented as optimizing:
\begin{gather}
\underset{s_1, ..., s_N}{\operatorname{argmax}} \sum_{t=1}^{N} \mathbb{E}_{A_t \sim p(A | s_{<t}, \mathcal{C}_t, \mathcal{Q}_t; \theta)}[U(Q_t, A_t; \theta)], \nonumber
\end{gather}
where $U(Q_t, A_t; \theta)$ is a \textbf{utility function}, implicitly learned or designed within the policy $\theta$, that evaluates the desirability of critiques $\mathcal{Q}_t$ to guide step selection.  This function allows the policy to interpret and leverage the nuanced information within NL critiques for enhanced reasoning.

\section{Experiment}
\label{sec:experiment}

\begin{table*}[t]
\centering
 \setlength{\tabcolsep}{9pt}
\begin{tabular}{l rrr rrrr} 
\toprule
\multirow{2}{*}{\textbf{Methods}} & \multicolumn{3}{c}{\bf AIME (Math)}    &   \multicolumn{4}{c}{\bf GPQA Diamond}\\ 
\cmidrule(lr){2-4} \cmidrule(lr){5-8}
 & \em 2024 & \em 2025  &   \em All   &  \em Biol.  & \em Chem.  &  \em Phys. &  \em All\\ 
\midrule
\multicolumn{8}{c}{\bf Llama3.1-8B-Instruct}\\
Baseline                        & 0.0 & 0.0 & 0.0 & 47.4 & 19.4 & 27.9 & 25.8 \\
\hdashline
Self-Consistency                & 3.3 & 0.0 & 2.2 & 36.8 & 28.0 & 31.4 & 30.3\\
Solution-Level Self-Evaluation  & 3.3 & 0.0 & 2.2 & 36.8 & 22.6 & 37.2 & 30.3 \\
~~~ + NL Self-Critique          & \em 13.3 & 0.0 & \textbf{8.9} & 36.8 & 26.9 & 37.6 & 32.5   \\
Step-Level Self-Evaluation      & 6.7 & 0.0 & 4.4 & \em 57.9 & 26.9 & 34.9 & 33.3 \\
\hdashline
\method (Ours)                  & 6.7 & 0.0 & 4.4 & 52.6 & \em 32.3 & \em 43.0 & \textbf{38.9} \\
\midrule
\multicolumn{8}{c}{\bf Llama3.3-70B-Instruct}\\
Baseline                        & 30.0 & 6.7 & 22.2 & 63.2 & 41.9 & 58.1 & 51.0\\
\hdashline
Self-Consistency                & \em 33.3 & 6.7 & \textbf{24.4} & 63.2 & \em 44.1 & 57.0 & 51.5\\
Solution-Level Self-Evaluation  & 26.7 & \em 13.3 & 22.2 & 63.2 & 40.9 & 61.6 & 52.0\\
~~~ + NL Self-Critique          & 26.7 & \em 13.3 & 22.2 & 63.2 & 40.9 & 62.8 & 52.5\\
Step-Level Self-Evaluation      & 23.3 & 6.7 & 17.8 & 63.2 & 39.8 & 61.6 & 51.5\\
\hdashline
\method (Ours)                  & 30.0 & \em 13.3 & \textbf{24.4} & 63.2 & 43.0 & \em 65.1 & \textbf{54.5}\\
\bottomrule
\end{tabular}
\caption{Experimental results on different reasoning tasks show that our method \method~outperforms both solution-level and step-level search algorithms by a significant margin. We highlight the best result in each individual domain in {\em italics} and the best overall result in {\bf bold}.  For a fair comparison, we use Self-Consistency with \( N = 5 \) examples. To better understand the impact of critique in search, we also present the results of using an external critique model.}
\label{tab:main}
\end{table*}

\subsection{Setup}

\paragraph{Benchmarks}
We conduct experiments on two benchmarks that assess the reasoning capabilities required for solving various scientific problems:
\begin{itemize}[leftmargin=10pt]
    \item \textbf{AIME}~\citep{aime}: a dataset from the American Invitational Mathematics Examination, which tests problem-solving skills across multiple areas of {\bf mathematics} (e.g., algebra, counting, geometry, and number theory). We include the two most recent test sets -- AIME2024 (30 problems) and AIME2025-Part1 (15 problems).
    \item \textbf{GPQA Diamond}~\cite{gpqa}: a challenging dataset of 198 multiple-choice questions written by domain experts in {\bf biology}, {\bf chemistry}, and {\bf physics}.
\end{itemize}

\paragraph{Baselines}
We compare our approach with two representative Self-Evaluation methods:
\begin{itemize}[leftmargin=10pt]
    \item \textbf{Self-Consistency} \citep{wangself}: This method first samples \( N \) reasoning paths and then selects the most consistent answer by marginalizing over the sampled reasoning paths.
    \item \textbf{Step-Level Self-Evaluation} \citep{xie2024self}: This method introduces a stepwise self-evaluation mechanism to guide and calibrate the reasoning process of LLMs. Specifically, it integrates self-evaluation guidance via stochastic beam search, facilitating an efficient search in the reasoning space.
\end{itemize}
In addition, we also consider two solution-level self-evaluation methods: 
\begin{itemize}[leftmargin=10pt]
    \item {\bf Solution-Level Self-Evaluation}: Instead of selecting the most consistent answer as done in Self-Consistency, this algorithm allows the model itself to select the final answer from the sampled \(N\) reasoning paths.
    \item {\bf Solution-Level Self-Evaluation with NL Self-Critique}: This method enhances the above method by incorporating a NL self-critique stage. After sampling \( N \) reasoning paths, the model generates a self-critique for each solution, assessing its correctness and plausibility. These self-critiques help the model to evaluate the quality of each reasoning path. The final answer is then selected based on this self-assessment process, aiming to improve the accuracy and reliability of the selected solution by promoting more informed decision-making.
\end{itemize}
We consider two backbone LLMs of different model sizes and capabilities, including Llama 3.1-8B-Instruct and Llama 3.3-70B-Instruct models. We generate 5 candidates at each step in the stepwise search process.

\subsection{Main Results}

\begin{figure*}
    \centering
    \includegraphics[width=\linewidth]{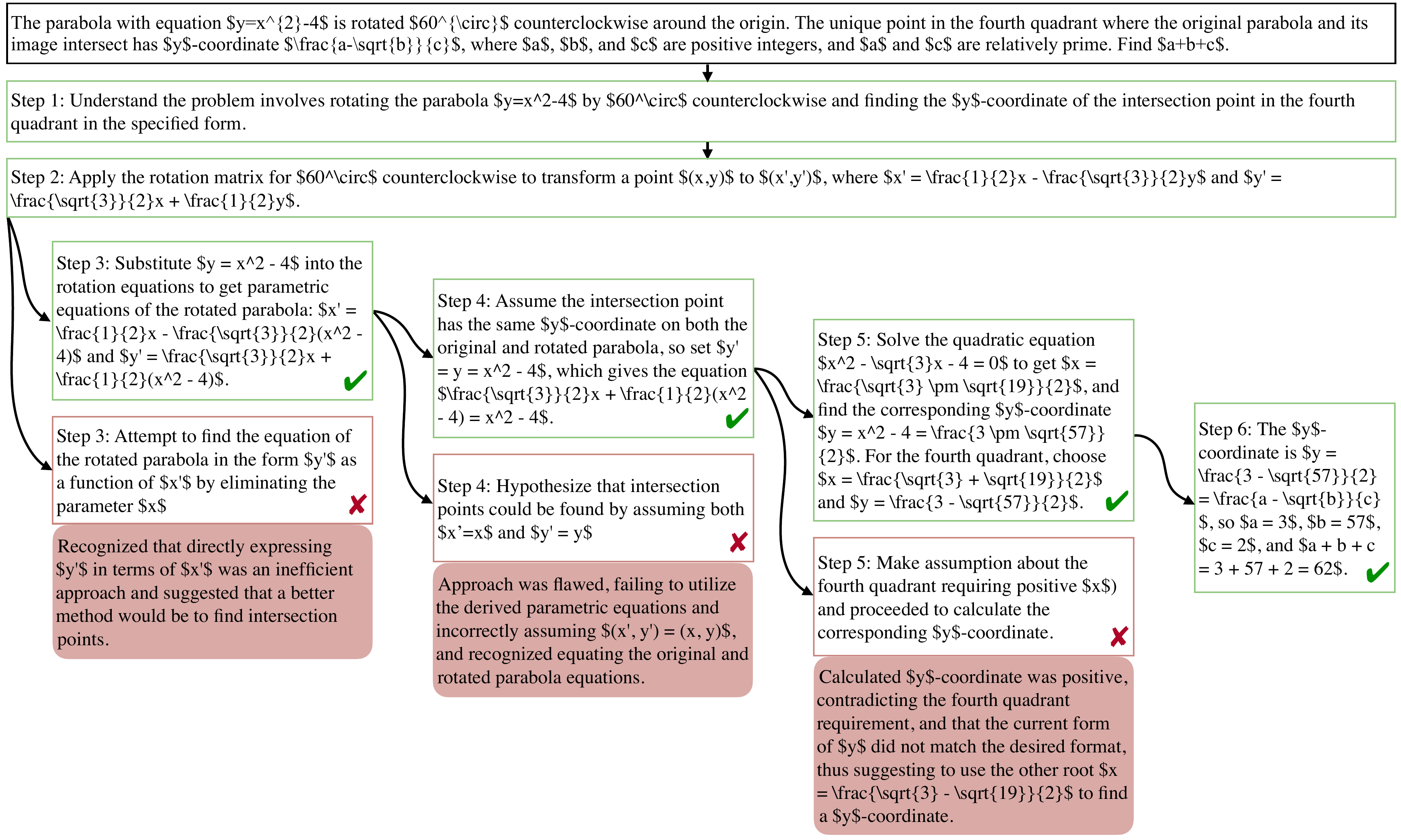}
    \caption{A case study from AIME25 where \method~produces correct results while step-level self-evaluation fails.}
    \label{fig:case}
\end{figure*}

Table~\ref{tab:main} presents the performance comparison of our proposed \method~framework against several baseline methods on the AIME and GPQA Diamond benchmarks, using two backbone LLMs: Llama3.1-8B-Instruct and Llama3.3-70B-Instruct. We have the following key observations.

\paragraph{\method~Outperforms Baselines Across Tasks and Model Sizes.}  
On the AIME-Math dataset, \method~achieves an accuracy of 4.4\% with the Llama3.1-8B-Instruct model, ranking the second-best methods in this setting. When utilizing the larger Llama3.3-70B-Instruct model, \method~improves the accuracy to 24.4\%, surpassing all other baselines. This demonstrates the effectiveness of \method~in enhancing the reasoning capabilities of LLMs, particularly as model size increases.

On the GPQA Diamond benchmark, which encompasses challenging questions from Biology, Chemistry, and Physics domains, \method~consistently achieves superior performance. With the Llama3.1-8B-Instruct model, \method~obtains the highest overall accuracy of 38.9\%, outperforming both solution-level and step-level self-evaluation methods. Notably, \method~achieves the best in the Chemistry (32.3\%) and Physics (43.0\%) domains. When using the larger Llama3.3-70B-Instruct model, \method~further improves the overall accuracy to 54.5\%, again surpassing all baselines and achieving the highest accuracy in the Physics domain (65.1\%). These results highlight \method's ability to handle intricate reasoning required in complex scientific domains.

\paragraph{Effectiveness of Natural Language Self-Critique.} 
The superior performance of \method~ can be attributed to its novel application of natural language self-critique as a feedback mechanism during the reasoning process. Introducing NL self-critique improves reasoning accuracy in both solution-level and step-level self-evaluation methods across different tasks and model sizes. Unlike traditional scalar reward verifiers used in self-evaluation methods, rich natural language critiques provide nuanced and interpretable feedback that guides the model toward more accurate reasoning paths. This approach enables the model to retain qualitative information about each reasoning step, directly addressing the limitations of existing scalar verifiers.
For instance, on the GPQA Diamond test set, NL self-critique improves reasoning accuracy over solution-level self-evaluation by 2.2\% for the Llama3.1-8B-Instruct model and by 0.5\% for the Llama3.3-70B-Instruct model. With the Llama3.3-70B-Instruct model, applying NL self-critique at the step level improves reasoning accuracy on the AIME and GPQA Diamond tasks by 6.6\% and 3.0\%, respectively.
Figure~\ref{fig:case} shows an example of how NL self-critique improves reasoning accuracy for the LLama3.3-70B-Instruct model.

\begin{figure*}[t!]
    \centering
    \subfloat[Llama3.1-8B-Instruct]{
    \includegraphics[width=0.35\linewidth]{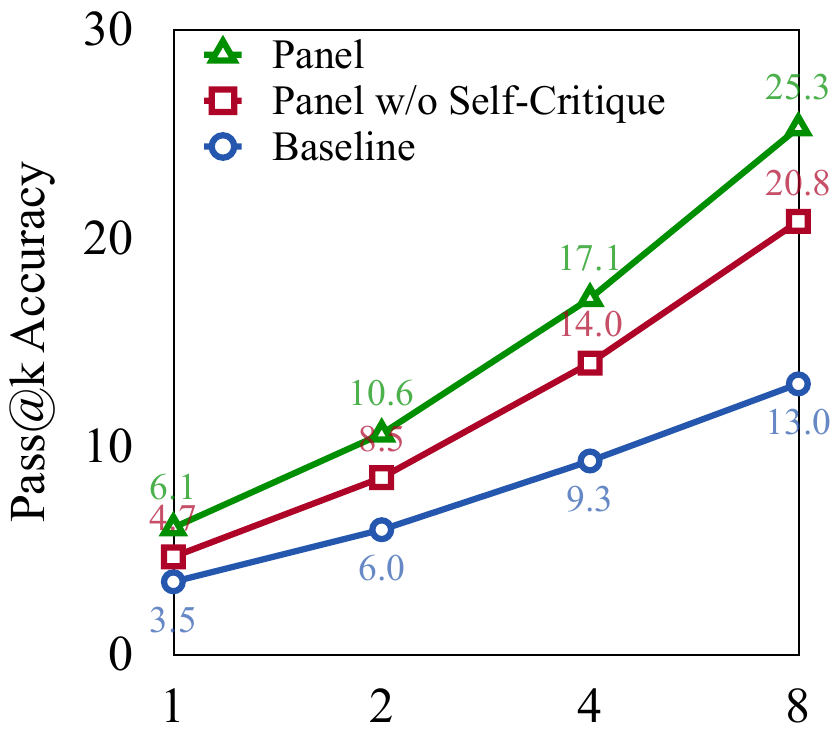}} \hspace{0.1\textwidth}
    \subfloat[Llama3.3-70B-Instruct]{
    \includegraphics[width=0.35\linewidth]{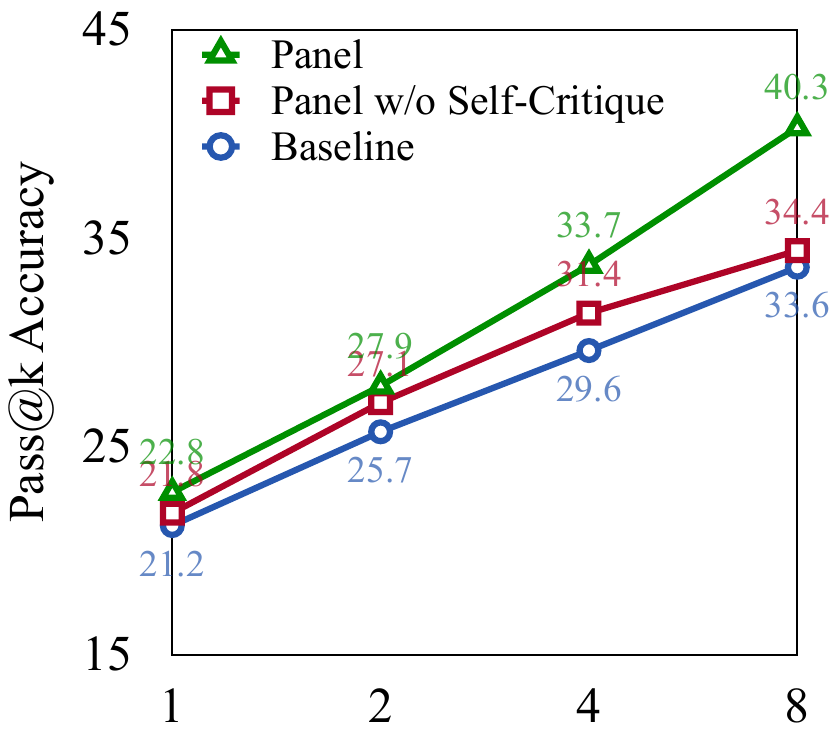}}
    \caption{Pass@k accuracies of our \method{} and the baseline model. For reference, we also provide the results of \method{} without NL Self-Critique (i.e., "Self-Level Self-Evaluation" in Table~\ref{tab:main}).}
    \label{fig:pass@k}
\end{figure*}


\subsection{Scalability of \method}

To further assess the scalability of \method{}, we examine the pass@k accuracy on the AIME2024-25 benchmark (see Figure~\ref{fig:pass@k}). This analysis explores how increasing the number of generated solutions \( k \) affects model performance, highlighting the trade-off between computational cost and accuracy.

Our findings show that \method{} consistently outperforms the baseline model across various values of \( k \) and model sizes. With the Llama3.1-8B-Instruct model, \method{} achieves a pass@1 accuracy of 6.1\%, nearly doubling the baseline's 3.5\%. As \( k \) increases, the performance gap widens; at pass@8, \method{} attains an accuracy of 25.3\%, significantly surpassing the baseline's 13\%. This indicates that \method{} is more effective at generating correct solutions when multiple attempts are allowed.

We also investigate the impact of NL self-critique by comparing \method{} with and without it. Removing the self-critique leads to a noticeable decline in accuracy across all \( k \) values and model sizes. For instance, with the Llama3.1-8B-Instruct model at pass@4, accuracy drops from 17.1\% with self-critique to 14.0\% without it. This demonstrates that NL self-critique provides valuable feedback that guides the model toward more accurate reasoning paths.
The benefits of NL self-critique are even more pronounced with larger models. Using the Llama3.3-70B-Instruct model at pass@8, incorporating self-critique boosts accuracy by 5.9\%, reaching 40.3\%. This suggests that larger models are better able to leverage the detailed feedback from self-critique to refine their reasoning processes.

In summary, the pass@k analysis illustrates that \method{} enhances the reasoning capabilities of LLMs, especially when generating multiple outputs at test time. By integrating NL self-critique, \method{} effectively improves accuracy while scaling with increased computational resources, aligning with our objective to promote more informed decision-making through qualitative feedback.

\subsection{Analysis}

In this section, we present a qualitative analysis to provide some insights into how \method{} improves reasoning accuracy.

\begin{table}[t]
\centering
\begin{tabular}{l rr} 
\toprule
\bf Critique    &   \bf AIME24-25   &   \bf GPQA\\
\midrule
\multicolumn{3}{c}{\bf Solution-Level Self-Evaluation}\\
Self (8B)       & 8.9 & 32.5\\
External (70B)  & \bf 11.1 & \bf 35.4\\
\midrule
\multicolumn{3}{c}{\bf Step-Level Self-Evaluation}\\
Self (8B)       & \bf 4.4 & \bf 38.9\\
External (70B)  & \bf 4.4 & 32.3\\
\bottomrule
\end{tabular}
\caption{Performance comparison of Llama3.1-8B-Instruct using self-critique versus an external larger (70B) critique model.}
\label{tab:critique}
\end{table}

\paragraph{NL Self-Critique is More Effective than Larger External Critique in Step-Level Self-Evaluation.}
We first investigate the impact of using an external critique model larger than the policy model on both solution-level and step-level self-evaluation methods.
As shown in Table~\ref{tab:critique}, employing an external 70B critique model improves performance in solution-level self-evaluation for both AIME24-25 (from 8.9\% to 11.1\%) and GPQA (from 32.5\% to 35.4\%). However, in step-level self-evaluation, the self-critique approach using the policy model itself (8B) outperforms the external critique on GPQA (38.9\% vs.\ 32.3\%) and matches performance on AIME24-25 (both at 4.4\%). This indicates that while larger external critique models can offer improvements at the solution level, the self-generated critiques from the policy model are more effective at refining reasoning steps, particularly in complex problem-solving tasks. The findings highlight the strength of our proposed \method~in leveraging self-critique to enhance reasoning without relying on larger external models.

\begin{figure}
    \centering
    \includegraphics[width=0.35\textwidth]{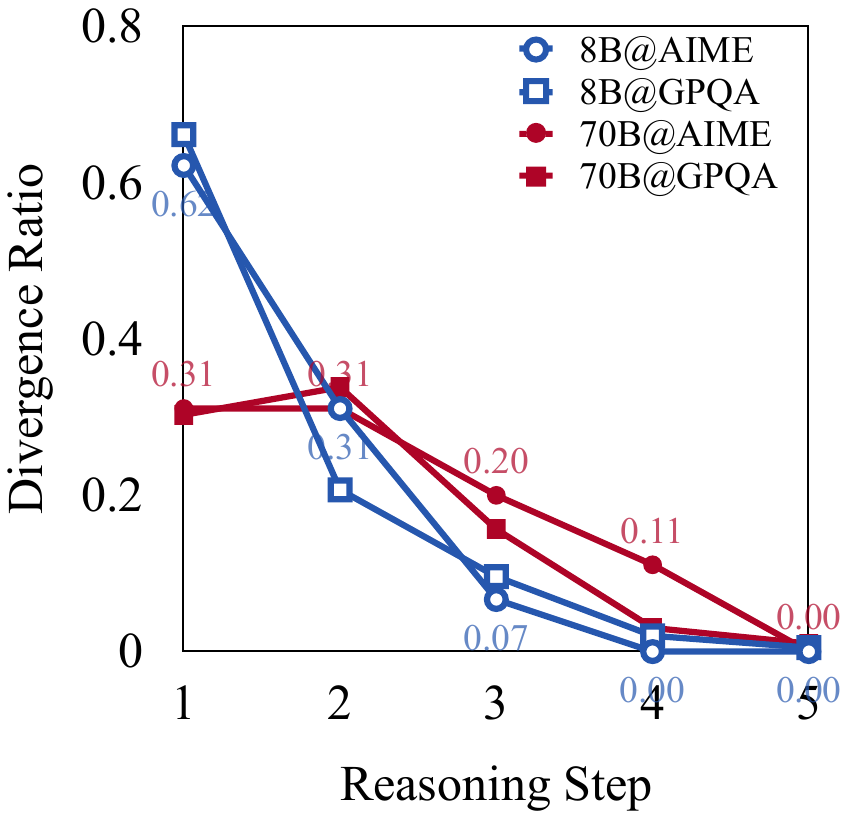}
    \caption{Impact of NL self-critique on decision making at each reasoning step. The "divergence ratio" denotes the proportion of decisions that differ when using NL self-critique versus not using it.}
    \label{fig:divergence}
\end{figure}

\paragraph{NL Self-Critique Significantly Influences Early Reasoning Steps.} 
Figure~\ref{fig:divergence} shows the impact of NL self-critique on decision making at each reasoning step.
Incorporating NL self-critique markedly affects the decision-making process of \method, particularly during the initial reasoning steps. The divergence ratio of decision making is highest at the first reasoning step across all models and benchmarks, and it decreases in subsequent steps.
For instance, when evaluating the Llama3.1-8B-Instruct model on the AIME benchmark, the divergence ratio at the first reasoning step is 62.2\%, indicating that more than half of the decisions differ due to the introduction of NL self-critique. This ratio decreases to 31.1\% at the second step and drops sharply to 6.7\% at the third step, suggesting that the influence of NL self-critique diminishes as the reasoning progresses. A similar pattern is observed on the GPQA benchmark, where the divergence ratio for the same model starts at 66.2\% and decreases to 0.5\% by the fifth step.
Similar trend can be found for the larger Llama3.3-70B model. 
These findings suggest that NL self-critique primarily influences the initial stages of the reasoning process, guiding the models toward more accurate or refined initial decisions. As the reasoning unfolds, the effects of self-critique become less pronounced, possibly because the initial decisions set the trajectory for subsequent steps. 

\section{Related Work}
\label{sec:related_work}

Our work is closely related to three key areas of research: inference time scaling, reward verification, and natural language critique. This section reviews recent advancements in these areas.

\paragraph{Inference Time Scaling} 
LLM reasoning extends to complex tasks such as logical inference \citep{kojima2022large, brown2020language}, step-by-step problem-solving \citep{wei2022chain}, and understanding cause-effect relationships \citep{wang2022rationale, zhouleast}. Reasoning structures like Tree-of-Thoughts \citep{yao2024tree, xie2024self} and Graph-of-Thoughts \citep{besta2024graph} incorporate meta-cognitive patterns like planning \citep{wang2023plan} and difficulty estimation \citep{fu2022complexity}.
Extending inference computation enhances reasoning abilities \citep{snell2024scaling}. Techniques such as Self-Consistency \citep{wangself, cobbe2021training} sample diverse reasoning paths to select the most consistent answers. Improved methods like boosted Self-Consistency \citep{pitis2023boosted} and increased sampling \citep{brown2024large, wu2024inference} enhance question coverage, yet selecting promising candidates remains challenging.

In this paper, we focus on step-level tree search to guide the LLM towards the promising reasoning trace. In contrast to solution-level scaling, our work introduces step-level algorithms with NL critique as the candidate selection strategy.

\paragraph{Reward Models and Verifiers}
Reward models and verifiers are crucial components in enhancing LLM reasoning. Traditionally, reward models primarily serve as learning signals for Reinforcement Learning (RL) of LLMs \citep{ouyang2022training, touvron2023llama}.  Building upon outcome-based rewards, Process Reward Models (PRMs) were introduced \citep{lightmanlet} to provide step-level feedback, demonstrably improving LLM reasoning performance.
The effectiveness of PRMs in step-level RL training has motivated their use as inference-time verifiers. One line of research focuses on training PRMs using policy rollout data and subsequently deploying them for online reasoning \citep{wang2024math, li2023making, hosseini2024v,lin2024critical}.  Another direction integrates PRMs as verifiers within Monte Carlo Tree Search (MCTS), aiming to mitigate the rollout overhead associated with MCTS frameworks \citep{luo2024improve, feng2023alphazero, tian2024toward}.

Unlike previous methods that use scalar scores for step evaluation, we leverage NL critiques as a novel reward signal. This fundamentally diverges from existing methodologies by moving beyond scalar feedback to leverage the rich information encoded in NL critiques.

\paragraph{Natural Language Critique}

NL critique, the process of generating natural language justifications for LLM reasoning, has emerged as a promising area.  Research demonstrates that LLM-generated critiques can effectively evaluate and refine the reasoning of other LLMs \citep{lin-etal-2024-criticbench}.  Consequently, NL critique has been leveraged to enhance agent performance across various tasks \citep{kim2023language, shinn2024reflexion, goucritic}, with some works focusing on training dedicated critique models \citep{cui2023ultrafeedback, ke2023critiquellm, ligenerative}.  These studies collectively highlight the potential of NL critique for Critique-Correcting Reasoning in diverse applications.

In contrast to this line of work, our paper explores the use of NL critiques and self-evaluation for inference-time search scaling at the step level.  While recent studies have integrated NL critique into search algorithms \citep{xie2024self, zhang2024generative}, their primary motivation is to improve verifier reward scores; Another study \citep{xi2024enhancing}, though utilizing step-level critique, mainly relies on an external strong LLMs to provide NL critique feedback. In contrast, we posit that NL self-critique is strong enough to guide policy search for self-improvement. This represents a fundamental departure from prior approaches.

\section{Conclusion}
\label{sec:conclusion}

In this work, we presented \method, a novel inference-time scaling framework that enhances LLMs reasoning by incorporating stepwise natural language self-critique into the step-level search process. Unlike traditional methods relying on scalar reward signals from external PRMs, \method~utilizes self-generated natural language critiques, providing rich, qualitative feedback essential for understanding and justifying complex reasoning steps. This approach addresses significant limitations of existing methods, such as the loss of nuanced information, the need for task-specific verifiers, and the associated computational overhead. Our experiments on challenging reasoning benchmarks demonstrate that \method~significantly outperforms traditional scalar reward-based methods, achieving substantial improvements in reasoning performance. 

By leveraging NL feedback, \method~opens new avenues for enhancing LLM reasoning capabilities across diverse tasks. Future work may explore further integration of NL feedback mechanisms and their applications in other domains.

\section*{Limitations}
While this work introduces a promising direction by leveraging natural language critiques for step-level inference scaling, limitations warrant consideration and future research. As a novel approach, this work represents an initial exploration of NL critique for step-level search.  While we provide empirical evidence supporting its effectiveness, a deeper theoretical understanding of why and when NL critiques are most beneficial is still needed. Moreover, quantifying and comparing information richness presents a methodological challenge.  While we posit that NL critiques offer richer step-level feedback compared to scalar reward values, establishing a direct quantitative comparison of this information richness is inherently difficult. Future research could investigate the information content and characteristics of effective NL critiques, and develop theoretical frameworks to better predict the performance gains achievable with this approach compared to existing methods.

\bibliography{ref}
\bibliographystyle{colm2024_conference}
\clearpage

\onecolumn
\appendix

\section{Prompt for NL self-critique in \method}
\label{sec:nl_critique_template}
We provide the prompts employed in \method~for NL self-critique, each specifically designed for distinct domains.

\begin{figure*}[htbp]
\centering
\begin{alprompt}{\centering {Math}}
You are an expert mathematician specializing in problem-solving and step-by-step reasoning. 
   Your task is to check the correctness of **the latest reasoning step** in solving a mathematical problem.\\

   **Important Instructions:**\\
   - Your goal is to determine if the current reasoning step contains any **logical, mathematical, or contextual errors**.\\
   - You should **focus on correctness**:\\
   - If the step is mathematically and logically valid given the context, mark it as "correct" regardless of whether it is incomplete or lacks further steps.\\
   - Do not penalize the step for not including subsequent steps unless its omission leads to a misunderstanding or error.\\

   **Common Errors to Look For:**\\
   1. Arithmetic or algebraic mistakes (e.g., incorrect simplifications or incorrect application of operations).\\
   2. Misapplied theorems or incorrect assumptions (e.g., an unjustified jump to a conclusion).\\
   3. Logical inconsistencies (e.g., a contradiction in the reasoning).\\
   4. Misinterpretation of the problem statement or prior steps.\\

   Based on the above guidelines, determine whether the current step is correct or incorrect. \\
   1. If it is correct, you should return "correctness": "correct" and "critique": "" (empty). \\
   2. If it is incorrect, you should return "correctness": "incorrect" and provide the explanation of the error in "critique". \\
   - Emphasize the core mistake(s) (location/reason).\\
   - **Keep it short and straightforward**. Avoid unnecessary detail.\\
   - If multiple errors, list them succinctly (e.g., bullet points).
\end{alprompt}
\caption{Prompt of  NL self-critique for math reasoning task.}
\end{figure*}

\begin{figure*}[b]
\centering
\begin{alprompt}{\centering {Physics}}
You are an expert physicist specializing in problem-solving and step-by-step reasoning across various subdomains of physics, including but not limited to Classical Mechanics, Electromagnetism, Quantum Mechanics, Thermodynamics, Relativistic Mechanics, Astrophysics, and Optics. 
   Your task is to check the correctness of **the latest reasoning step** in solving a physics problem. \\

   **Important Instructions:** \\
   - Your goal is to determine if the current reasoning step contains any **logical, mathematical, or conceptual errors** specific to physics. \\
   - You should **focus on correctness**: \\
   - If the step is physically and logically valid given the context, mark it as "correct" regardless of whether it is incomplete or lacks further steps. \\
   - Do not penalize the step for not including subsequent steps unless its omission leads to a misunderstanding or error. \\

   **Common Errors to Look For:** \\
   1. **Mathematical Errors**: \\
      - Incorrect calculations, algebraic manipulations, or unit conversions. \\
      - Misapplication of formulas or equations. \\
   2. **Conceptual Errors**: \\
      - Misinterpretation of physical laws or principles (e.g., Newton's laws, conservation of energy, or Maxwell's equations). \\
      - Incorrect assumptions or simplifications. \\
   3. **Logical Errors**: \\
      - Contradictions in the reasoning or unjustified jumps to conclusions. \\
      - Misinterpretation of the problem statement or prior steps. \\
   
   **Examples of Critique:** \\
   - "The calculation of the force is incorrect because Newton's second law was misapplied." \\
   - "The energy conservation principle was violated in this step, leading to an incorrect result." \\
   - "The reasoning assumes a constant velocity, which contradicts the problem's context." \\

   Based on the above guidelines, determine whether the current step is correct or incorrect. \\
   - If it is correct, you should return "correctness": "correct" and "critique": "" (empty).  \\
   - If it is incorrect, you should return "correctness": "incorrect" and provide the explanation of the error in "critique". \\
      - Emphasize the core mistake(s) (location/reason). \\
      - **Keep it short and straightforward**. Avoid unnecessary detail. \\
      - If multiple errors, list them succinctly (e.g., bullet points). \\
\end{alprompt}
\caption{Prompt of NL critique for Physics task.}
\end{figure*}

\begin{figure*}[htbp]
\centering
\begin{alprompt}{\centering {Chemistry}}
You are an expert chemist specializing in problem-solving and step-by-step reasoning across various subdomains of chemistry, including but not limited to Organic Chemistry, Inorganic Chemistry, Physical Chemistry, and Analytical Chemistry. 
   Your task is to check the correctness of **the latest reasoning step** in solving a chemistry problem. \\

   **Important Instructions:** \\
   - Your goal is to determine if the current reasoning step contains any **logical, mathematical, or conceptual errors** specific to chemistry. \\
   - You should **focus on correctness**: \\
   - If the step is chemically and logically valid given the context, mark it as "correct" regardless of whether it is incomplete or lacks further steps. \\
   - Do not penalize the step for not including subsequent steps unless its omission leads to a misunderstanding or error. \\

   **Common Errors to Look For:** \\
   1. **Mathematical Errors**: \\
      - Incorrect calculations, stoichiometric ratios, or unit conversions. \\
      - Misapplication of formulas or equations (e.g., ideal gas law, equilibrium constants). \\
   2. **Conceptual Errors**: \\
      - Misinterpretation of chemical principles or laws (e.g., Le Chatelier's principle, reaction mechanisms, or periodic trends). \\
      - Incorrect assumptions or simplifications (e.g., ignoring side reactions or assuming ideal behavior). \\
   3. **Logical Errors**: \\
      - Contradictions in the reasoning or unjustified jumps to conclusions. \\
      - Misinterpretation of the problem statement or prior steps. \\

   **Examples of Critique:** \\
   - "The stoichiometric calculation is incorrect because the mole ratio was misapplied." \\
   - "The reaction mechanism violates the conservation of mass due to an unbalanced equation." \\
   - "The reasoning assumes ideal gas behavior, which contradicts the problem's context of high pressure." \\

   Based on the above guidelines, determine whether the current step is correct or incorrect. \\
   - If it is correct, you should return "correctness": "correct" and "critique": "" (empty). \\
   - If it is incorrect, you should return "correctness": "incorrect" and provide the explanation of the error in "critique".  \\
      - Emphasize the core mistake(s) (location/reason). \\
      - **Keep it short and straightforward**. Avoid unnecessary detail. \\
      - If multiple errors, list them succinctly (e.g., bullet points). \\

\end{alprompt}
\caption{Prompt of  NL self-critique for chemistry task.}
\end{figure*}

\begin{figure*}[htpb]
\centering
\begin{alprompt}{\centering {Biology}}

You are an expert biologist specializing in problem-solving and step-by-step reasoning across various areas of biology. 
   Your task is to check the correctness of **the latest reasoning step** in solving a biology problem. \\

   **Important Instructions:** \\
   - Your goal is to determine if the current reasoning step contains any **logical, factual, or conceptual errors** specific to biology. \\
   - You should **focus on correctness**: \\
   - If the step is biologically and logically valid given the context, mark it as "correct" regardless of whether it is incomplete or lacks further steps. \\
   - Do not penalize the step for not including subsequent steps unless its omission leads to a misunderstanding or error. \\

   **Common Errors to Look For:** \\
   1. **Factual Errors**: \\
      - Incorrect use of biological facts, terminology, or definitions (e.g., confusing mitosis with meiosis or misidentifying biomolecules). \\
      - Misinterpretation of experimental data or observations. \\
   2. **Conceptual Errors**: \\
      - Misapplication of biological principles or theories (e.g., natural selection, central dogma, or Mendelian inheritance). \\
      - Incorrect assumptions or simplifications (e.g., ignoring environmental factors or assuming ideal conditions). \\
   3. **Logical Errors**: \\
      - Contradictions in the reasoning or unjustified jumps to conclusions. \\
      - Misinterpretation of the problem statement or prior steps. \\
   4. **Mathematical Errors**: \\
      - Incorrect calculations or statistical analyses (e.g., error in population genetics or enzyme kinetics). \\

   **Examples of Critique:** \\
   - "The reasoning incorrectly assumes that all mutations are harmful, which contradicts the concept of neutral mutations." \\
   - "The calculation of the allele frequency is incorrect because the Hardy-Weinberg equilibrium conditions were not met." \\
   - "The interpretation of the experimental results ignores the possibility of confounding variables."

\end{alprompt}
\caption{Prompt of  NL self-critique for biology task.}
\end{figure*}

\clearpage
\section{Case Study of \method~on STEM Task}
\begin{figure*}[h]
    \centering
    \includegraphics[width=\linewidth]{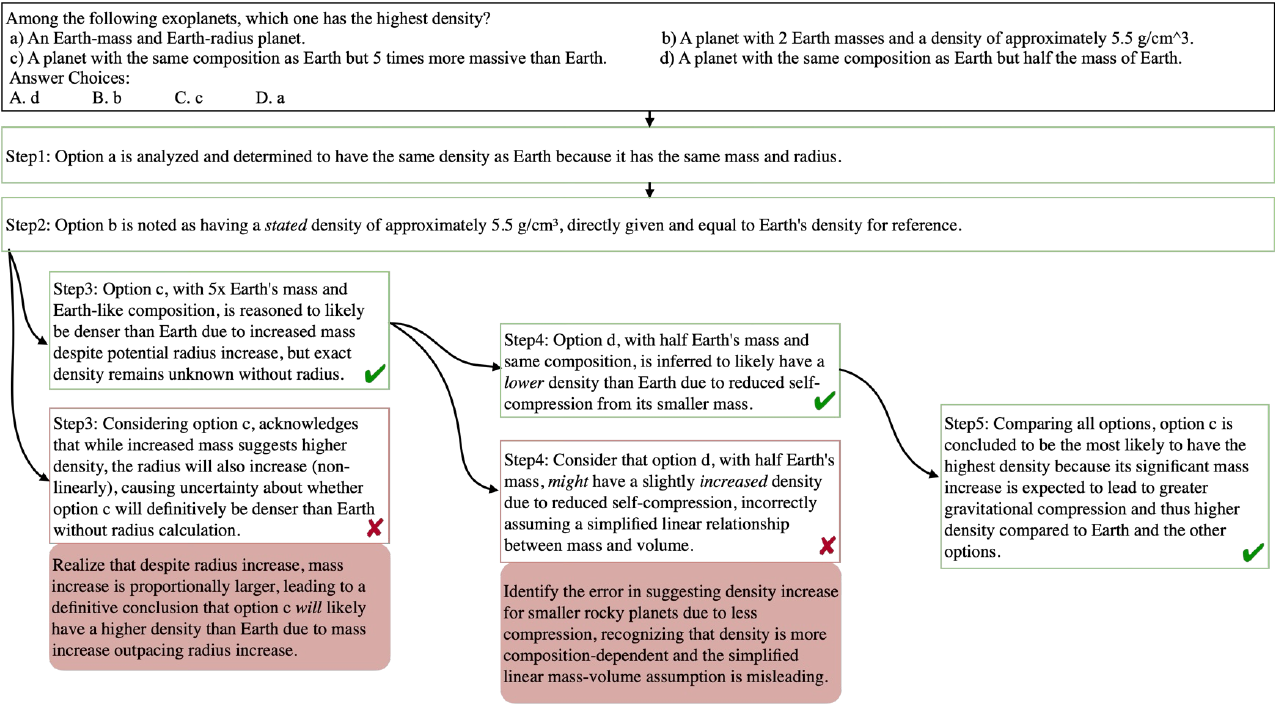}
    \caption{A case study from GPQA Diamond where \method~ produces the correct reasoning trace while step-level self-evaluation fails.}
    \label{fig:case_physics}
\end{figure*}

\paragraph{Example of \method~Inference-Time Search}
We also provide a case study of our \method~inference-time search examples in STEM fields (i.e. GPQA Diamond in Physics) in Figure~\ref{fig:case_physics}.

\end{document}